%% file: main.tex
\documentclass[]{bytedance}
\usepackage[toc,page,header]{appendix}

\usepackage{minitoc}
\usepackage{amsfonts}
\usepackage{amssymb}
\usepackage{tabularx}
\usepackage{listings}
\usepackage{xcolor}
\usepackage{cancel}

\usepackage{tabulary,multirow,xspace}
\usepackage{fixmath,mathtools,nicefrac,mmstyle}
\usepackage{subcaption}
\usepackage{caption}
\usepackage{wrapfig} 
\usepackage[misc]{ifsym} 
\usepackage{colortbl}

\usepackage{wrapfig}
\usepackage{multicol}
\usepackage[most]{tcolorbox}
\usepackage{pifont}

\definecolor{codegreen}{rgb}{0,0.6,0}
\definecolor{codegray}{rgb}{0.5,0.5,0.5}
\definecolor{codepurple}{rgb}{0.58,0,0.82}
\definecolor{backcolour}{rgb}{0.95,0.95,0.92}
\definecolor{boxblue}{RGB}{57,89,163}
\definecolor{boxbluebg}{RGB}{230,237,250} 

\lstdefinestyle{mystyle}{
    backgroundcolor=\color{backcolour},   
    commentstyle=\color{codegreen},
    keywordstyle=\color{magenta},
    numberstyle=\tiny\color{codegray},
    stringstyle=\color{codepurple},
    basicstyle=\ttfamily\footnotesize,
    breakatwhitespace=false,         
    breaklines=true,                 
    captionpos=b,                    
    keepspaces=true,                 
    numbers=none,                    
    numbersep=5pt,                  
    showspaces=false,                
    showstringspaces=false,
    showtabs=false,                  
    tabsize=2
}
\definecolor{mygray1}{gray}{.95}
\definecolor{mygray2}{gray}{.9}
\definecolor{mygray3}{gray}{.95}
\usepackage{pifont}

\newcommand{\myparagraph}[1]{{\noindent\bf #1}}

\newlength\savewidth
\newcolumntype{x}[1]{>{\centering\arraybackslash}p{#1pt}}

\newcommand{\app}{\raise.17ex\hbox{$\scriptstyle\sim$}}

\makeatletter
\DeclareRobustCommand\onedot{\futurelet\@let@token\@onedot}
\def\@onedot{\ifx\@let@token.\else.\null\fi\xspace}

\def\eg{\emph{e.g}\onedot} 
\def\ie{\emph{i.e}\onedot}

\makeatother

\makeatletter

\newcommand{\Rmnum}[1]{\expandafter\@slowromancap\romannumeral #1@}
\makeatother

\usepackage{xcolor}
\input{macros}

\usepackage{listings}

\definecolor{commentgreen}{rgb}{0.1, 0.4, 0.1}
\definecolor{keywordblue}{rgb}{0.1, 0.1, 0.7}
\definecolor{stringred}{rgb}{0.7, 0.1, 0.1}

\lstdefinestyle{mystyle}{
    commentstyle=\color{commentgreen},
    keywordstyle=\color{keywordblue},   
    stringstyle=\color{stringred},
    basicstyle=\ttfamily\scriptsize, 
    breaklines=true,
    keepspaces=true,
    showstringspaces=false,
    frame=none,                     
    language=Python, 
}

\newcommand{\name}{DreamPoster}
\title{\name{}: A Unified Framework
for Image-Conditioned Generative Poster Design}

\author{
\centerline{
    Xiwei Hu $^{1,*}$\quad 
    Haokun Chen $^{1,*}$ \quad  
    Zhongqi Qi $^{1,*}$ \quad 
    Hui Zhang $^{1,2,*}$ \quad
    \vspace{5pt}
} 
\centerline{
    Dexiang Hong $^{1}$ \quad
    Jie Shao $^{1,{\dagger}}$\quad
    Xinglong Wu $^{1}$ \quad 
    \vspace{-5pt}
}
}

\affiliation[1]{Intelligent Creation Lab, ByteDance}
\affiliation[2]{Fudan University}
\contribution[*]{Equal contribution}
\contribution[\dagger]{Project lead}

\abstract{
We present DreamPoster, a Text-to-Image generation framework that intelligently synthesizes high-quality posters from user-provided images and text prompts while maintaining content fidelity and supporting flexible resolution and layout outputs. Specifically, DreamPoster is built upon our T2I model, Seedream3.0~\cite{gao2025seedream} to uniformly process different poster generating types. For dataset construction, we propose a systematic data annotation pipeline that precisely annotates textual content and typographic hierarchy information within poster images, while employing comprehensive methodologies to construct paired datasets comprising source materials (e.g., raw graphics/text) and their corresponding final poster outputs. Additionally, we implement a progressive training strategy that enables the model to hierarchically acquire multi-task generation capabilities while maintaining high-quality generation. Evaluations on our testing benchmarks demonstrate DreamPoster's superiority over existing methods, achieving a high usability rate of 88.55\%, compared to GPT-4o~\cite{gpt4o20250325} (47.56\%) and SeedEdit3.0~\cite{wang2025seededit} (25.96\%). DreamPoster will be online in Jimeng and other Bytedance Apps.
}

\date{\today}

\checkdata[Project Page]{\url{https://dreamposter.github.io/}}

\correspondence{Xiwei Hu, Jie Shao at \email{\{huxiwei, shaojie.mail\}@bytedance.com}}

\begin{document}
\maketitle

\input{sec/intro}

\input{sec/related_work}

\input{sec/dataset}
\input{sec/methods}

\input{sec/exps}

\input{sec/conclusion}

\clearpage

\bibliographystyle{plainnat}
\bibliography{main}

\end{document}

%% file: macros.tex
\usepackage{graphicx}
\usepackage{amssymb}
\usepackage{pifont}
\usepackage{floatrow}
\usepackage{amsmath} 
\usepackage{float}
\usepackage{wrapfig}
\usepackage{multirow}
\usepackage{tcolorbox}
\tcbuselibrary{breakable, skins, raster}
\usepackage{listings}

%% file: sec/intro.tex
\section{Introduction}\label{sec:intro}
\begin{figure}[htbp] %
\vspace{-1.0em}
     \centering
     \includegraphics[width=0.8\textwidth]{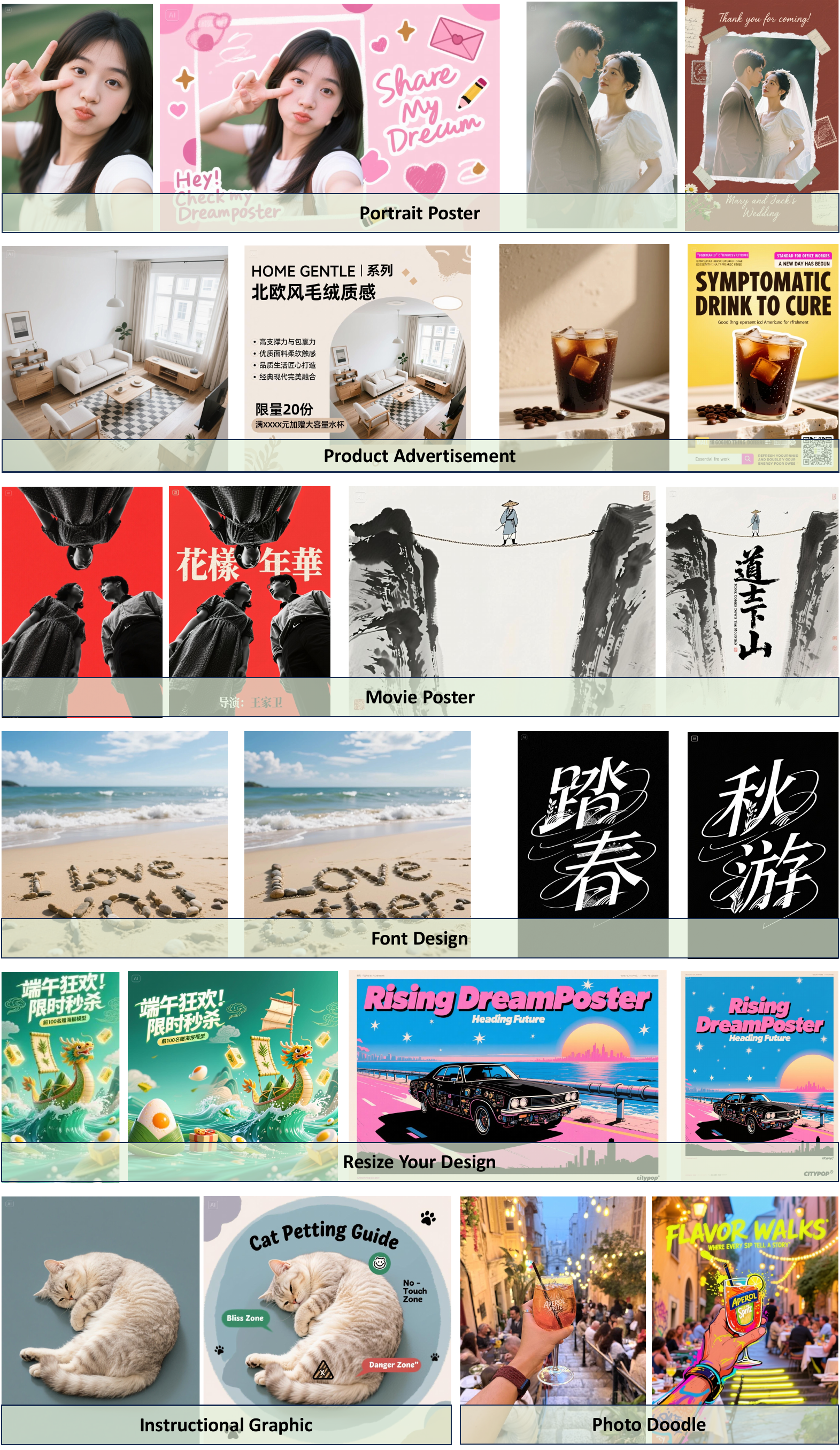}
     \caption{A wide variety of posters generated by DreamPoster. Note that all input images are generated by Seedream3.0~\cite{gao2025seedream}}
    \label{fig:teaser}
    \vspace{-0.5em}
\end{figure}

Recent advancements in Text-to-Image (T2I) generation have yielded significant progress, facilitating breakthroughs across diverse domains ranging from artistic creation to advertising design. However, models that rely exclusively on text prompts to guide image generation often struggle to meet the nuanced and fine-grained demands of real-world creative tasks, particularly when users seek to build upon pre-existing visual content. In practical design contexts such as the creation of promotional posters, advertisements, and social media covers, there is a growing need for Image-to-Image (I2I) generation capabilities that seamlessly integrate user-provided images with textual directives, resulting in cohesive, high-quality visuals. To effectively address these practical requirements, models must extend beyond mere textual comprehension and possess the ability to interpret and transform image content while maintaining both fidelity and design coherence.

Despite the progress made in generative models, existing Image-to-Image methods, such as image editing~\cite{brooks2023instructpix2pix,huang2024smartedit,feng2025dit4edit,lin2024schedule,zhang2023magicbrush,zhao2024ultraedit} and unified image generation~\cite{gpt4o20250325, gemini25flash2025, mou2025dreamo, xiao2024omnigen}, show several limitations. Prior methods struggle to fully integrate multimodal information, resulting in outputs that lack subject preservation, aesthetic appeal, or layout flexibility. For example, models such as Step1X-Edit~\cite{liu2025step1x} and SeedEdit1.6~\cite{shi2024seededit} enforce rigid input-output aspect ratio constraints, restricting their adaptability and limiting their practical utility. Models like GPT-4o~\cite{gpt4o20250325}, while powerful, often produce unstable results in terms of aspect ratio and subject preservation, frequently yielding distorted or poorly composed outputs. Additionally, the design sense in its generated content tends to be suboptimal, often lacking visual harmony or balance. To this end, we present DreamPoster, a unified text-to-image generation framework for image-conditioned poster design. 
DreamPoster generates high-quality posters from user-provided inputs. The inputs include a base image, such as a product photo or illustration, and text prompts like a title, tagline, or description. The model integrates the visual and textual content into a coherent and visually appealing layout.
Unlike general-purpose generators, DreamPoster is specifically trained to maintain content fidelity while producing an aesthetically pleasing layout. Our approach supports flexible resolutions and aspect ratios, addressing the limitation of prior systems that forced fixed dimensions. DreamPoster is built upon a powerful diffusion-based T2I model (Seedream3.0~\cite{gao2025seedream}) and extends it with a multimodal architecture and a curated training strategy for the poster domain.

The contributions of this work are threefold. First, we construct a large-scale dataset of posters paired with their source components using a novel data curation pipeline. This includes automated filtering for text visibility and aesthetics, and a custom Poster Captioner that annotates each poster with detailed textual and layout descriptors. These annotations enable learning of fine-grained typographic hierarchy and design attributes. Second, we design a unified generative architecture that concatenates text tokens, image embeddings, and noise in a transformer-based diffusion model, allowing DreamPoster to jointly model layout, visual, and textual information. We fine-tune the base diffusion model with a progressive multi-stage training regimen: starting with simple text addition, then multi-task mixed editing, and finally high-quality fine-tuning for aesthetic alignment. Third, we conduct comprehensive evaluations against state-of-the-art baselines. DreamPoster significantly outperforms competing methods in human evaluations of prompt following, subject preservation, and design sense. In the usability study, our framework achieved an 88.55\% success rate, far higher than GPT-4o (47.56\%) and our earlier SeedEdit3.0 model (25.96\%). DreamPoster will be deployed in real-world content creation applications (\eg ByteDance’s Jimeng platform), enabling users to generate professional-grade posters from minimal inputs.

%% file: sec/related_work.tex
\section{Related Work}\label{sec:related}

\subsection{Image-to-Image Generation}
Recent advances in Text-to-Image (T2I) models have enabled the generation of high-quality images from textual descriptions. However, T2I models often struggle to meet the fine-grained control and content preservation requirements of real-world creative tasks. 
To address this limitation, Image-to-Image (I2I) generation tasks have emerged, where the model is required to follow both the input image and the prompt, transforming the image into a user-desired version while still preserving the key content or structure of the original image.
Depending on the specific application scenario and intent, I2I tasks can be categorized into various types, including but not limited to: image editing~\cite{brooks2023instructpix2pix,huang2024smartedit,feng2025dit4edit,lin2024schedule,zhang2023magicbrush,zhao2024ultraedit}, subject-driven generation~\cite{zhang2025icedit,mou2025dreamo,zhang2025easycontrol,ruiz2023dreambooth,cai2024dsd,tan2024ominicontrol,shin2024diptych,chen2024anydoor}, style transfer~\cite{wang2024instantstyle,xu2024freetuner,wang2023stylediffusion,zhang2023inversion}, and so on.

Traditionally, most methods have designed and trained specialized expert models for different I2I sub-tasks, resulting in limited flexibility in multi-task scenarios. Recently, some unified approaches~\cite{shi2024seededit,wang2025seededit,gpt4o20250325,liu2025step1x,gemini25flash2025,labs2025fluxkontext} have been proposed to cover multiple I2I tasks within a single framework. For example, models such as GPT-4o~\cite{gpt4o20250325}, Gemini2.5Flash~\cite{gemini25flash2025}, the SeedEdit series~\cite{shi2024seededit,wang2025seededit}, and Step1X-Edit~\cite{liu2025step1x} integrate multimodal inputs (\eg, image and text) within a unified generative architecture, achieving cross-task image generation and editing capabilities. 

\subsection{Intelligent Poster Generation}
Poster design demands not only seamless integration of visual and textual elements, but also imposes higher requirements on layout hierarchy, typographic style, and overall aesthetic harmony. Early automatic poster generation systems~\cite{lin2023autoposter,jia2023cole} were primarily based on rule-driven templates, resulting in limited creative expression and diversity constrained by the templates themselves.
With the rapid advancement of diffusion models and large-scale multimodal models, some recent works~\cite{gao2025postermaker,wang2025designdiffusion,chen2025posta,pu2025art,zhang2025creatidesign,zhang2024creatilayout,liu2024glyphbyt5,chen2023textdiffuser2,tuo2024anytext2} have attempted to build end-to-end unified frameworks for automated poster generation. However, these methods often struggle with overall aesthetic coordination and precise adherence to design elements, and are typically limited to handling only a single sub-task of I2I generation(\ie poster generation), lacking support for diversified editing operations or multi-style adaptation.

To this end, we propose DreamPoster, which leverages a unified diffusion transformer architecture to enable intelligent poster generation across multiple tasks, styles, and layouts. Our approach not only ensures content fidelity and layout flexibility, but also significantly enhances the design quality and aesthetic consistency of generated posters, better satisfying the diverse and high-standard requirements of real-world applications.

%% file: sec/dataset.tex
\section{Dataset Pipeline}\label{sec:dataset}
In this section, we describe the systematic pipeline used to curate the DreamPoster dataset. Our pipeline involves data filtering, source-target pairing, and detailed re-captioning, ensuring each sample includes clear text, strong aesthetics, and fine-grained layout annotations. This provides high-quality, structured data for subsequent model training.

\begin{figure}[h]
  \centering
  \includegraphics[width=1\linewidth]{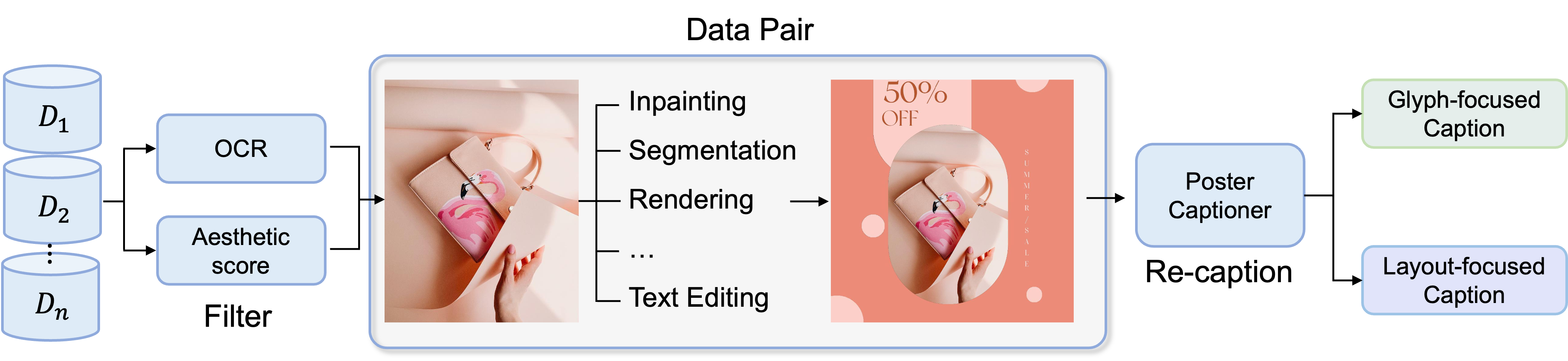}
  \caption{The overview of our dataset pipeline, which consists of data filtering, data pair construction and re-captioning. }
  \label{fig:dataset_pipeline}
\end{figure}

\subsection{Data Filtering} 
Training a generative model for posters requires paired data of source content (images and text) and the final designed poster. However, such paired data is not readily available. We assemble a large dataset of existing high-quality posters and devise an automated pipeline to extract source materials and annotations for each. To ensure data quality, we first filter a collection of candidate poster images using OCR and aesthetic scoring. We discard images where OCR cannot recognize any text and those with low aesthetic appeal. Only images above a certain aesthetic score and with clear textual content are retained as our clean poster dataset.

\subsection{Data pair Construction}  
To construct paired source-target training examples, we create for each poster an equivalent “deconstructed” source input. Using a suite of advanced image processing techniques, we reverse-engineer each final poster into its constituent parts. Specifically, we apply context-aware inpainting to remove textual content from the poster image, producing a clean background image that retains the main graphic elements but without any overlaid text. We also perform semantic segmentation to isolate key visual components (such as the primary product or figure in the poster) from the background if needed. The original text from the poster is saved as the source text prompt. In some cases, we retrieve or approximate the exact font and style of the text and render it separately. Through these diverse methods, we ensure the training pairs have structural correspondence while preserving semantic fidelity – the model sees how a given image and pieces of text can be combined to achieve the final designed poster. Overall, our data pipeline yields a large number of training pairs covering diverse poster styles and layouts, all annotated with fine-grained captions for auxiliary guidance.

\subsection{Poster Captioner} We generate rich annotations for each poster. We train a Poster Captioner, a specialized image captioning model for text-heavy images. It is able to identify all text strings present and additionally describes their visual attributes such as font style, size, color, and arrangement. The Poster Captioner produces two complementary captions for every poster: \Rmnum{1}) a glyph-focused caption detailing fine-grained typographic properties (\eg ``Large bold red title text `Summer Sale' at top, smaller white subtitle below in cursive font.''), and \Rmnum{2}) a layout-focused caption describing the overall spatial arrangement (\eg ``A poster featuring a toy as the main subject. The main title reads `Cute Toy,' and the subtitle says `Buy One Get One Free.' The composition is symmetrical.''). By recaptioning each poster in this way, we provide the generative model with supervisory signals to learn both the visual appearance of glyphs and the hierarchical layout of design elements. 

%% file: sec/methods.tex
\section{Architecture and Training Strategy}\label{sec:method}

\subsection{Model Architecture}
DreamPoster is built on a diffusion-based image generation backbone that integrates both image and text conditions. We adopt a transformer-based diffusion architecture, following recent Diffusion Transformer (DiT) designs that replace the usual UNet with a transformer for improved scalability~\cite{flux1dev,gao2025seedream}. In our implementation, we treat all inputs as a single sequence of tokens: this sequence includes tokens encoding the conditional image (the input image), tokens encoding the text prompt, and tokens for the noisy target image latent. Respective positional embeddings are added so that the model can distinguish the roles of each token. By concatenating text and image embeddings along the sequence dimension, the model can attend to both modalities jointly through self-attention layers, achieving a seamless fusion of visual and textual information. This design is inspired by prior works~\cite{tan2024ominicontrol} that demonstrated the effectiveness of transformer backbones for multimodal generation, where image patches and text tokens are processed together to allow fine-grained alignment between text and image content. We initialize our model with a strong pre-trained text-to-image diffusion model (Seedream 3.0~\cite{gao2025seedream}) and then adapt it to the poster domain. During training, most of the transformer layers are initialized from the pre-trained model’s weights, but we fine-tune specific portions of the network to specialize in our task. 

\begin{figure}[h]
  \centering
  \includegraphics[width=1\linewidth]{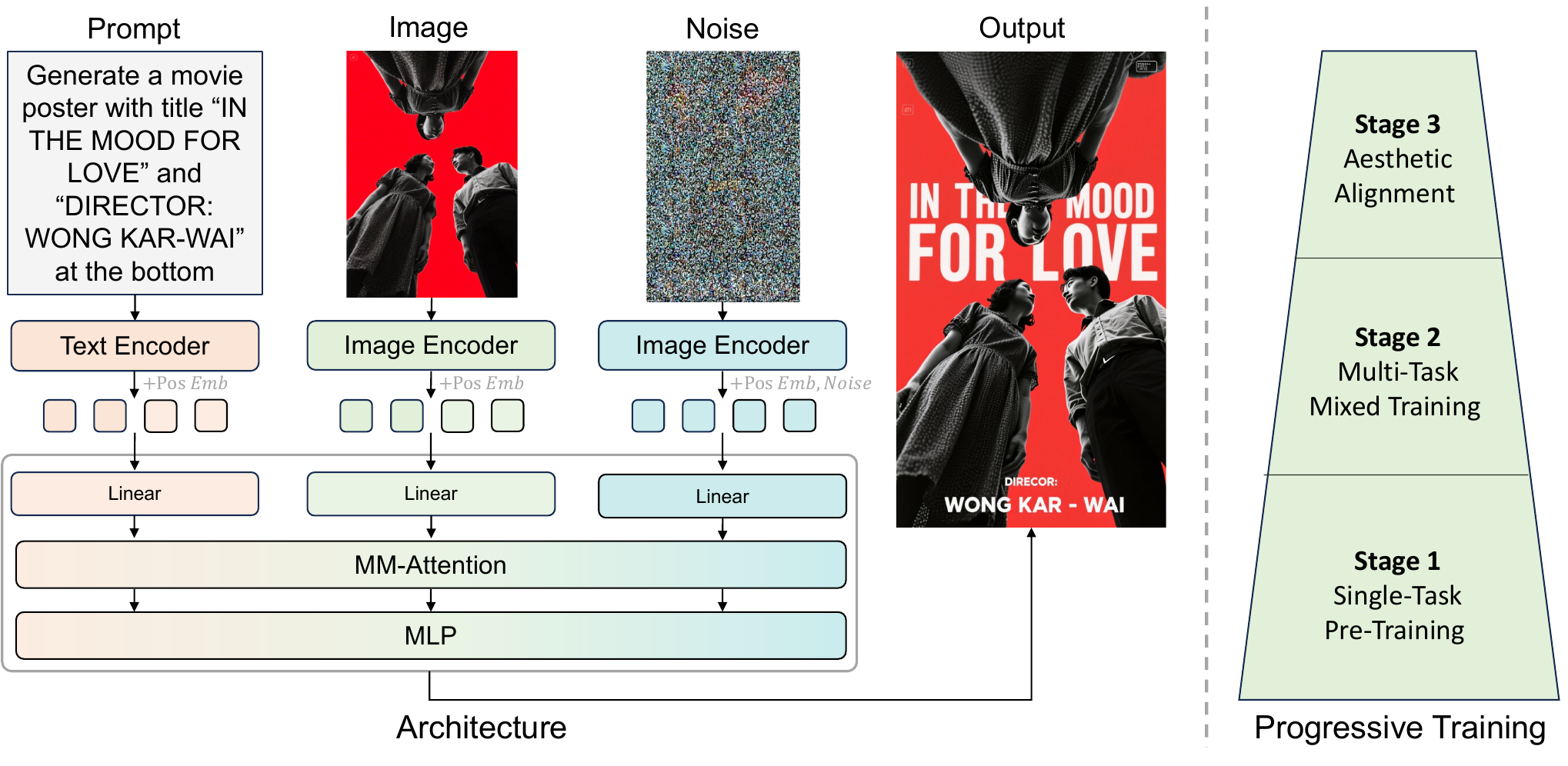}
  \caption{{\bf The overview of our proposed DreamPoster.} {\it Left:} We concatenate condition image embedding and text embedding to generate various posters with the given image. {\it Right:} The progressive training strategy of DreamPoster. }
  \label{fig:arch}
\end{figure}

\subsection{Progressive Training Strategy}

We train DreamPoster using a three-stage curriculum designed to gradually endow the model with more complex design abilities. Each stage introduces new training data and objectives, building on the skills learned in the previous stage:

\myparagraph{Stage 1: Single-Task Pretraining (Text Addition).}
 In the first stage, we focus on the simplest task: adding text to an image. The model is trained on a subset of data where the goal is to place given text onto the input image in a reasonable location and style, without significantly altering the image otherwise. This task teaches the model the basic alignment between textual content and background images, essentially learning to “write” on images.

\myparagraph{Stage 2: Multi-Task Mixed Training.}
In the second stage, we expose the model to a mixture of different poster generation tasks. In addition to text addition, the training data now includes samples of text modification, text deletion, multi-aspect design, poster restyling, and other variations. We shuffle and mix these tasks during training so that the model learns to handle diverse and complex editing instructions. This multi-task training expands the model’s capabilities beyond simple caption placement – it learns to, for example, replace old text with new text seamlessly, adjust layouts, or apply stylistic changes as directed by the prompt. By the end of this stage, DreamPoster has a broad skill set for poster design operations, all under a unified framework.

\myparagraph{Stage 3: Fine-Grained Aesthetic Alignment.}
The final stage finetunes the model on a small high-quality dataset with expert-designed posters, focusing on the subtleties of professional design. Here we train the model to optimize for visual aesthetic quality and layout refinement. 
After this stage, DreamPoster can produce outputs that are not only correct in content but also visually appealing and coherent at a professional level.

Through these progressive stages, the training effectively follows a curriculum from easy tasks to hard tasks. This strategy allows the model to maintain high generation quality while expanding its functionality. By Stage 3, DreamPoster has been optimized to handle complex design compositions and fine aesthetic details, achieving a high level of creative autonomy in poster generation.

%% file: sec/exps.tex
\section{Experiments}\label{sec:exp}

\begin{figure}[h]
  \centering
  \includegraphics[width=1\linewidth]{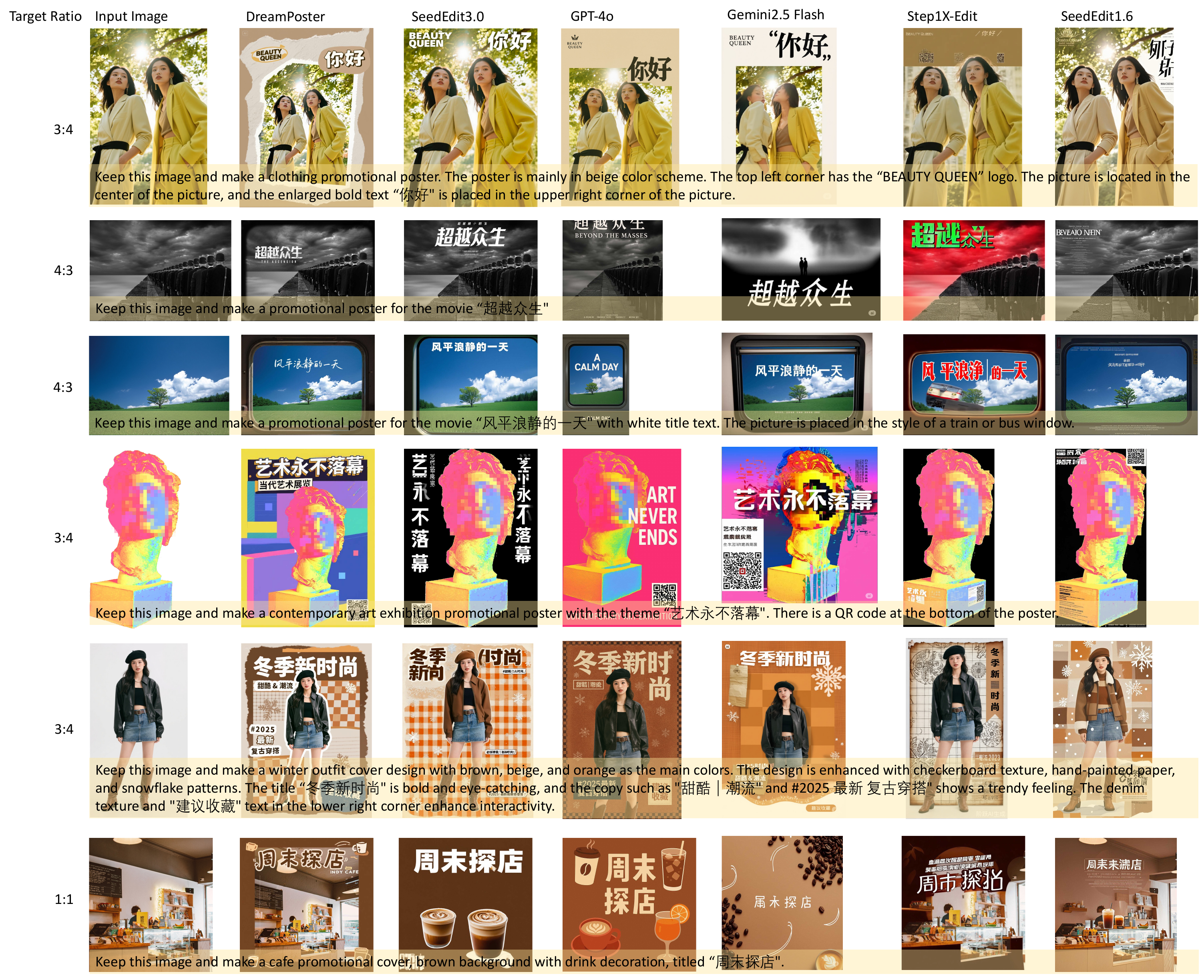}
  \caption{Qualitative comparisons between DreamPoster and other methods. Notice the advantage of DreamPoster in prompt following, subject preservation, and design sense. Note that GPT-4o and Gemini2.5 exhibit instability in target aspect ratio preservation during image synthesis, while Step1X-Edit and SeedEdit1.6 enforce rigid input-output dimensional alignment through fixed aspect ratio constraints.}
  \label{fig:quanlitative_result}
\end{figure}

\subsection{Qualitative Results}
As illustrated in Figure~\ref{fig:quanlitative_result}, we show representative qualitative comparisons between DreamPoster and several state-of-the-art poster generation models, including SeedEdit3.0~\cite{wang2025seededit}, GPT-4o~\cite{gpt4o20250325}, Gemini2.5Flash~\cite{gemini25flash2025}, Step1X-Edit~\cite{liu2025step1x}, and SeedEdit1.6~\cite{shi2024seededit}. These examples cover a variety of poster types such as product promotion, event advertising, and creative artwork. Compared to all baselines, DreamPoster demonstrates consistently superior alignment between user prompts and generated outputs. In particular, it exhibits more accurate rendering of user-specified texts (\eg main title, slogan), better spatial integration with input images, and stronger control over typographic hierarchy. In particular, DreamPoster is capable of generating diverse layouts that preserve the visual prominence of the subject while maintaining clean, professional aesthetics.

In contrast, it has been observed that models such as GPT-4o and Gemini2.5Flash frequently struggle to maintain the intended aspect ratio or introduce layout artifacts, particularly when aligning multiple textual elements against complex backgrounds. While Step1X-Edit and SeedEdit1.6 enforce stringent input-output dimensional constraints, their inability to generate flexible layouts results in visually rigid outputs that often lack balance. Furthermore, SeedEdit3.0 predominantly treats text as simple overlays, which undermines overall design coherence. In contrast, DreamPoster effectively learns to reconcile content, typography, and layout through a progressive training process coupled with rich design supervision. These qualitative observations substantiate our quantitative findings, demonstrating that DreamPoster achieves a superior trade-off between prompt fidelity, visual consistency, and design quality across a diverse array of real-world scenarios.

\begin{figure}[h]
  \centering
  \includegraphics[width=0.9\linewidth]{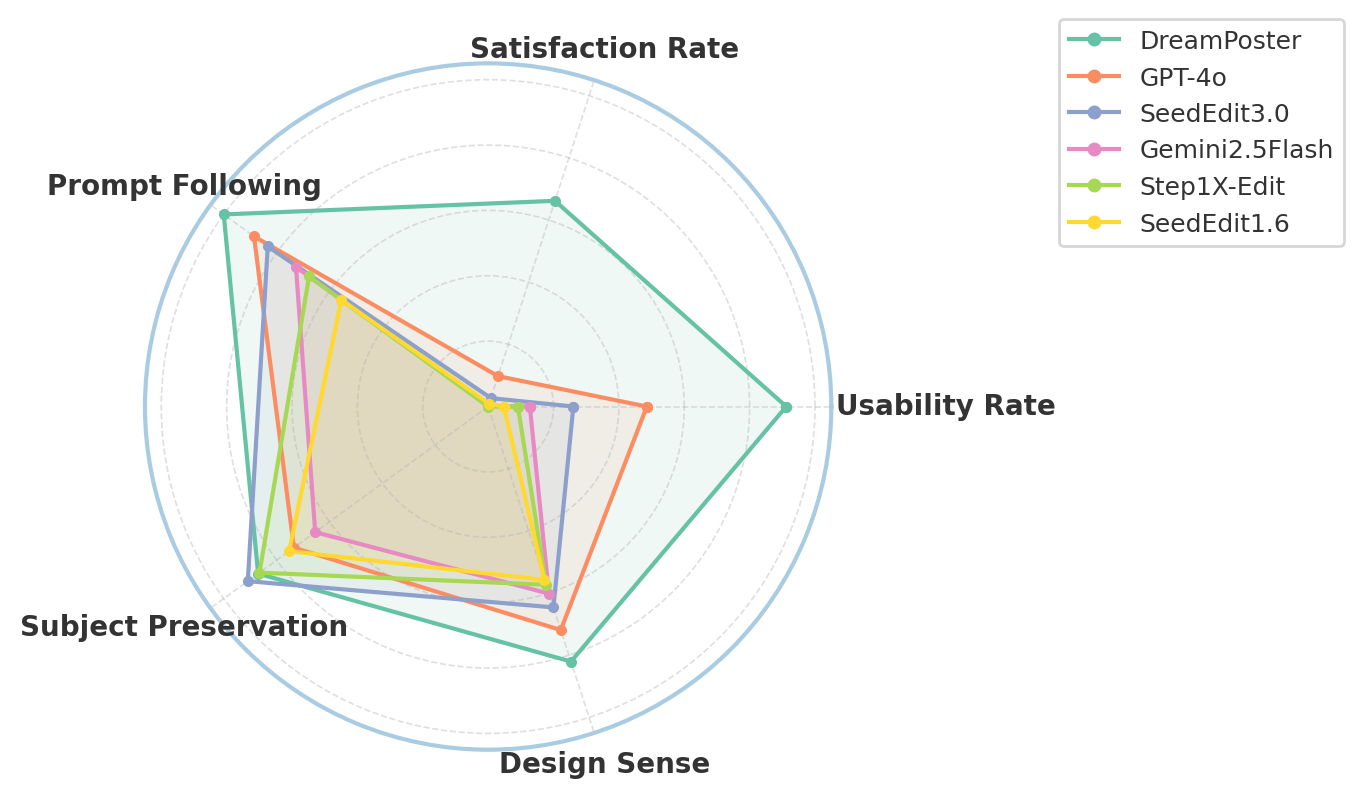}
  \caption{Weight normalized radar chart comparing DreamPoster with baseline models on five evaluation metrics (higher is better for all). DreamPoster (red) achieves the highest scores across all metrics, indicating a superior balance of instruction compliance, content preservation, and design quality, which translates into dramatically higher usability and user satisfaction.}
  \label{fig:ai_model_radar_chart}
\end{figure}

\subsection{Quantitative Results}
We conduct a comprehensive evaluation of DreamPoster using an internally curated benchmark featuring diverse poster design scenarios, including movie posters, product advertisements, holiday cards, and event promotions. This dataset is designed to assess whether the model can effectively integrate user-provided images and prompts into well-formed, aesthetically pleasing posters. We define three expert evaluation criteria—Prompt Following, Subject Preservation, and Design Sense—to assess whether the model (1) adheres to input instructions, (2) preserves the main content of the provided image, and (3) produces coherent and professional layout designs. In addition, we conduct a large-scale user study involving 40 participants (30 with graphic design experience), who evaluate the outputs based on the Usability Rate (the percentage of results with fewer than 3 non-satisfaction points) and the Satisfaction Rate (the percentage of fully satisfactory results with no issues).

As shown in Fig~\ref{fig:ai_model_radar_chart}, DreamPoster consistently outperforms all baseline models across all five evaluation dimensions. It achieves an 88.55\% Usability Rate, significantly surpassing GPT-4o (47.56\%) and SeedEdit3.0 (25.96\%). In expert-rated dimensions, DreamPoster attains the highest scores in Prompt Following (3.88/5.00) and Design Sense (3.19/5.00), while maintaining strong Subject Preservation (3.38/5.00). These results demonstrate that DreamPoster not only faithfully follows user instructions but also excels in visual fidelity and design quality, leading to significantly higher user satisfaction and real-world usability compared to all baselines.

%% file: sec/conclusion.tex
\section{Conclusion}\label{sec:con}
In this report, we present DreamPoster, a unified framework for image-conditioned generative poster design that substantially advances the state-of-the-art in AI-driven poster design. We constructed a high-quality dataset through a novel pipeline that gathers and refines image-poster pairs, ensuring rich and diverse training data for the task. Building on this foundation, we developed a unified generative model with multimodal fusion, seamlessly integrating visual and textual inputs. By gradually fine-tuning the model through stages of increasing task complexity, we achieve stable learning and enhanced creative fidelity in the generated posters. Extensive experiments and evaluations confirm the effectiveness of DreamPoster. The model delivers significant improvements in all key metrics, outperforming strong baselines such as GPT-4o and SeedEdit3.0 by a wide margin.